\pdfoutput=1

\documentclass[11pt]{article}

\usepackage[final]{acl}

\usepackage{times}
\usepackage{latexsym}

\usepackage[T1]{fontenc}
\usepackage{tracefnt}

\usepackage[utf8]{inputenc}

\usepackage{microtype}

\usepackage{inconsolata}

\usepackage[disable]{todonotes}   
\usepackage{array}
\usepackage{makecell}
\usepackage{multirow}
\usepackage{mathtools}
\usepackage{enumitem}
\usepackage{amssymb}

\usepackage{graphicx}
\usepackage[dvipsnames]{xcolor}
\usepackage{algorithm}
\usepackage{algpseudocode}

\newcommand{\ACRO}[1]{\textsc{#1}}

\newcommand{\CRACS}{\ACRO{CODI/CRAC}}

\title{Improving LLMs' Learning of Coreference Resolution}
\title{Improving LLMs' Learning for Coreference Resolution}

\author{
  Yujian Gan\textsuperscript{1} \quad
  Yuan Liang\textsuperscript{1} \quad
  Yanni Lin\textsuperscript{2} \quad
  Juntao Yu\textsuperscript{1} \quad
  Massimo Poesio\textsuperscript{1,3} \\
  \textsuperscript{1}Queen Mary University of London \\
  \textsuperscript{2}Guangxi Normal University \\
  \textsuperscript{3}University of Utrecht \\
  \texttt{y.gan@qmul.ac.uk} \quad \texttt{yuan.liang@qmul.ac.uk} \quad
  \texttt{linyn@mailbox.gxnu.edu.cn} \quad \\
  \texttt{juntao.yu@qmul.ac.uk} \quad
  \texttt{m.poesio@qmul.ac.uk} \\
}

\begin{document}
\maketitle
\begin{abstract}

Coreference Resolution (CR) is crucial for many NLP tasks, but existing LLMs struggle with hallucination and under-performance.
In this paper, we investigate the limitations of existing LLM-based approaches to CR—specifically the Question-Answering (QA) Template and Document Template methods—and propose two novel techniques: Reversed Training with Joint Inference and Iterative Document Generation.
Our experiments show that Reversed Training improves the QA Template method, while Iterative Document Generation eliminates hallucinations in the generated source text and boosts coreference resolution.
Integrating these methods and techniques offers an effective and robust solution to LLM-based coreference resolution~\footnote{Our code is available \href{https://github.com/ygan/LLMs-Coreference}{here}.}.

\end{abstract}

\section{Introduction}

Coreference resolution involves detecting and clustering different mentions that refer to the same discourse world entity.
As a task that requires linguistic and extra-linguistic understanding, it plays a crucial role for many downstream natural language processing tasks, such as information extraction, text summarization, chatbots, and dialogue systems.
As a result, CR has garnered significant attention from the NLP community~\cite{poesio2023computational}.

The evolution of coreference resolution models can be divided from rule-based and statistical learning approaches to deep learning approaches ~\citep{liu2023brief, poesio2023computational}.
In the past few years, Large Language Models (LLMs), which implicitly incorporate contextual information and commonsense knowledge, have revolutionized NLP by significantly improving performance across many tasks.
This advancement has led researchers to investigate LLMs' potential for coreference resolution.

Recent studies have already proved the feasibility of prompting LLMs to resolve coreferences, with a special focus on zero- and few-shot applications \citep{yang-etal-2022-gpt, agrawal-etal-2022-large, le2023largelanguagemodelsrobust, zhu-etal-2024-large, gan-etal-2024-assessing}.
\citet{le2023largelanguagemodelsrobust} have shown that prompt-based LLMs surpass previous unsupervised systems but still perform worse than state-of-the-art supervised models. In their experiments, two prompt templates were used, namely the Question-Answering (QA) Template and the Document Template, as shown in Figure~\ref{fig:QA-document-template-example}.
Specifically, \citet{le2023largelanguagemodelsrobust} have focused more on the Document Template, while \citet{gan-etal-2024-assessing} have concentrated on the QA Template.
Experimental results in our comparative study show that the performance of the Document Template is superior to that of the QA Template.

However, it is found that the Document Template method relies on a key assumption: LLMs do not generate hallucinations when resolving coreferences. 
But LLMs do produce hallucinations, which is challenging to match the generated document with the correct places in the original text. 
For example, in Figure~\ref{fig:QA-document-template-example}, if the LLM wrongly generates ``There are a candle a wall ...'' as ``There are a candle a candle a wall ...'' with an unnecessary mention ``a candle'' repeated, it becomes much harder to align these mentions with the original document.
Through studying the code given by \citet{le2023largelanguagemodelsrobust}, we have noticed that they removed examples where LLMs created hallucinations, only keeping cases where the original text matched correctly (i.e., without hallucinations).
However, we have found that hallucinations are quite common, especially in linguistically complex texts, where the problem becomes even more noticeable.
Although the QA Template method is less affected by hallucinations, its performance is weaker than the Document Template method, making it less reliable for coreference resolution.

To overcome the limitations of both the QA and Document Template methods, we propose two new approaches: Reversed Training with Joint Inference, and Iterative Document Generation.
Our experiments show that the Iterative Document Generation method not only completely removes hallucinations of LLMs but also improves the final CoNLL score. 
Meanwhile, the Reversed Training with Joint Inference method significantly improves the LLM’s CR performance when using the QA Template. 
Ablation studies also show that removing any component of this solution reduces the CoNLL score.

Overall, our work makes two key contributions:

\begin{figure*}[h]
    \includegraphics[width=0.95\linewidth]{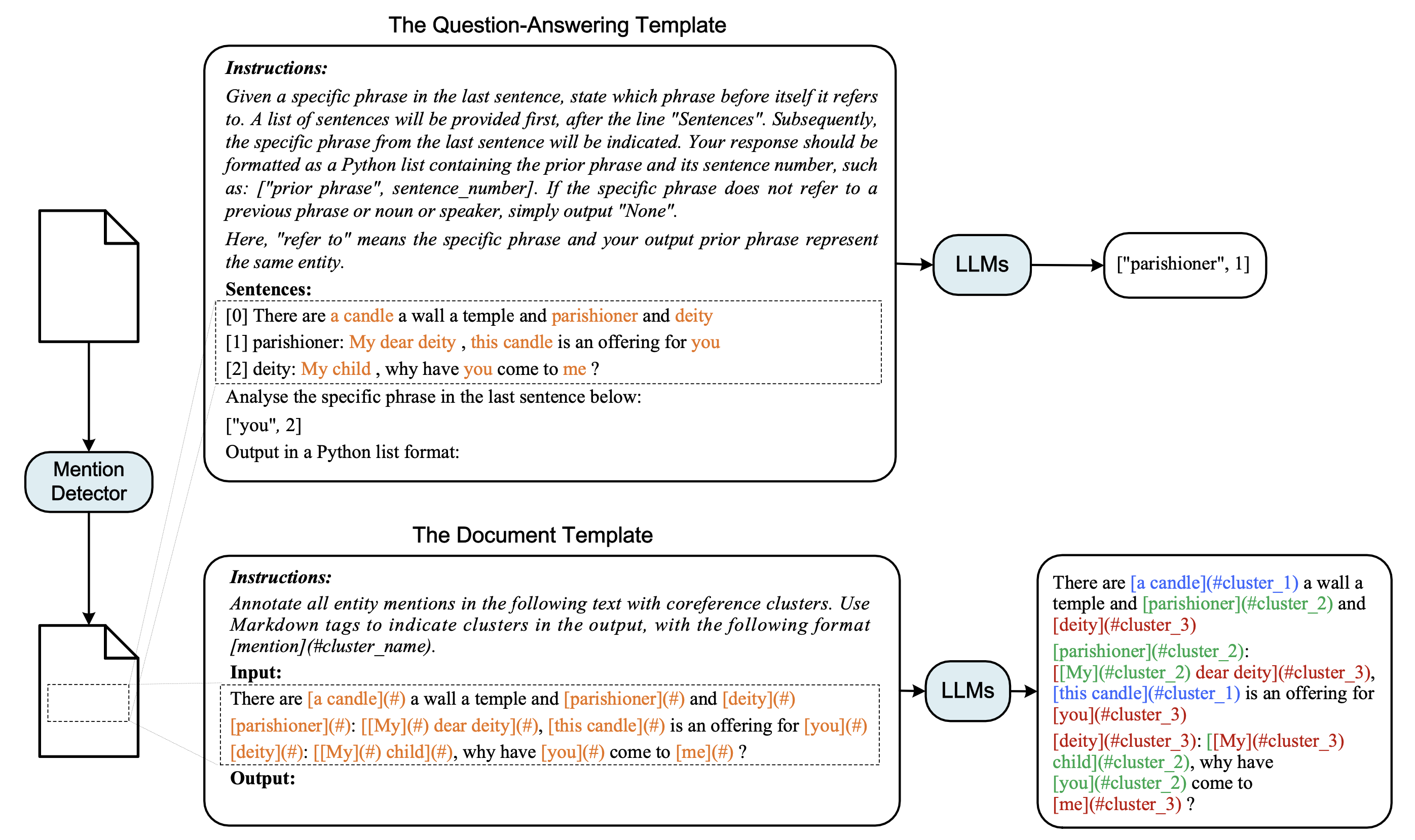}

  \caption{An example of the QA and Document Templates for Coreference Resolution. In the QA Template, LLMs generate answers based on input content and questions. In the Document Template, LLMs analyze the marked mention and assign a cluster ID to it.}
  \label{fig:QA-document-template-example}
  \vspace{-1em}
\end{figure*}

\begin{enumerate}

    \item We propose the Reversed Training and Joint Inference methods for coreference resolution based on the QA Template. Ablation study shows that these methods effectively improve LLM performance.

    \item  We propose the Iterative Document Generation method for coreference resolution based on the Document Template. Experiments demonstrate that this method not only enhances LLM performance, but also effectively tackles the issue of hallucination in LLMs when using the Document Template.

\end{enumerate}


\section{Related Work}
\subsection{Traditional Coreference Resolution Systems}
Typically, traditional coreference resolution systems
have addressed the task through mention-ranking models.
Early approaches, such as those by \citet{wiseman-etal-2015-learning} and \citet{clark-manning-2016-improving}, used a two-step process. 
First, a mention detector identified potential mentions in the text.
Then, a separate clustering step grouped these mentions into coreference clusters.
\citet{lee-etal-2017-end} introduced a pivotal advancement with a deep learning system that jointly performed mention detection and clustering.
This end-to-end approach proved to be both effective and simple, becoming the standard architecture for coreference resolution for a considerable period.

Building on this foundation, subsequent work explored various enhancements.
\citet{lee-etal-2018-higher} and \citet{kantor-globerson-2019-coreference} incorporated contextual embeddings to improve performance.
\newcite{yu-etal-2020-cluster} focused on addressing the challenges posed by singletons and non-referring expressions.
\newcite{joshi-etal-2019-bert,joshi-etal-2020-spanbert} further refined the model by replacing LSTMs with fine-tuned BERT and SpanBERT, harnessing the power of transformer architectures.

While these advancements refined the core architecture, research has also began to shift towards exploring alternative approaches, such as reformulating the task using question-answering frameworks and document annotation templates.

\subsection{Coreference Resolution as Question Answering}
One line of studies have investigated the potential of framing coreference resolution as a question-answering task.

\newcite{wu-etal-2020-corefqa} recast coreference resolution as the task of finding all other mentions (answers) that belong to the same cluster as a given mention (question). They utilized SpanBERT, pre-trained on Quoref and SQuAD 2.0, to encode the document and employed a BIO scheme to tag valid answers for each mention.  Their findings highlighted the importance of considering scores from both directions (i.e., $S_{i,j}$ and $S_{j,i}$ for a mention pair \textit{i}, \textit{j}), which resonates with our observations on reversed training discussed later in this paper (Section \ref{subsec:reversed-training}).
\newcite{aralikatte-etal-2021-ellipsis} adopted a BERT-based machine reading comprehension (MRC) approach, focusing on resolving ellipsis by identifying the answer span (antecedent) for a given mention.
\newcite{yang-etal-2022-gpt} evaluated early GPT models (e.g., GPT-2) on the ECB+ corpus in a few-shot setting, using a QA Template for binary judgments on gold mention pairs. However, this early attempt did not achieve significant results.
\newcite{bohnet-etal-2023-coreference} achieved state-of-the-art results by fine-tuning the mT5-XXL (13B) model to output candidate mentions and their corresponding clusters.

More recently, \citet{gan-etal-2024-assessing} have demonstrated prompt-based instruction-tuned language models was feasible to resolve coreference. One of the two templates used in their study was the Question-Answering (QA) Template, where the language model generates the answer when given a passage and an open-ended \textit{wh}-question.
With further exploration on this QA Template, \citet{gan-etal-2024-assessing} have conducted a comprehensive evaluation on the capabilities of different LLMs to resolve coreferences. In \citet{gan-etal-2024-assessing}'s QA Template, the antecedent and its sentence ID was generated in a pair.
In the same year, \newcite{zhu-etal-2024-large} have conducted a similar evaluation of several recent LLMs on prompt-based coreference resolution as a component of context understanding.
A key difference from \citet{gan-etal-2024-assessing}'s work is that \newcite{zhu-etal-2024-large} formulated the task as a multiple-choice problem.

\subsection{Document Annotation-based Coreference Resolution}
Another line of research have explored the use of document annotation for coreference resolution.

\citet{zhang-etal-2023-seq2seq} introduced a seq2seq approach using a prompt-fine-tuned T0 model. This system took a document as input and produces the same document with annotated mentions and clusters. Their results demonstrated that this approach can achieve performance comparable to that of \citet{bohnet-etal-2023-coreference}. \citet{le2023largelanguagemodelsrobust} have proposed a prompt-based method where LLMs were provided with documents containing highlighted mentions (either predicted or gold) followed by a special cluster placeholder (\#). The LLMs were then tasked with predicting and filling the placeholders with appropriate cluster IDs.
However, our preliminary experiments with this method revealed issues including hallucination. Our solutions to this problem will be given in Section \ref{sec:Document-Template}.

Our work mainly builds upon the approach of \citet{le2023largelanguagemodelsrobust} and \citet{gan-etal-2024-assessing}, further refining the QA and Document Template to achieve significant improvements.

\subsection{Reverse Training}
The ``Reversal Curse'' describes LLMs' inability to understand the symmetric property of identify relation, i.e., LLMs trained on ``A is B'' may struggle to learn ``B is A'' \citep{berglund2024reversalcursellmstrained}.

The Reversal Curse has been remedied by the recent attempts of reverse training.
\citet{Yu2024ReverseMI} have demonstrated that LLMs can learn from modeling in both directions with comparable proficiency. In their study, a reverse training sample was constructed by directly reversing the original token sequence.
\citet{golovneva2024reversetrainingnursereversal} have proved the effectiveness of training by reversing the ordering of segmented text chunks.
In addition to positional relationship in linear sequences, the concept of ``reverse'' can be generalized to more types of association in many cases, where information presented in the reverse order is no less valuable than that in the original order.
In multi-modal tasks, for example, \citet{gul-artzi-2024-cogen} have reversed the input of image description and the output of image classification.

Inspired by the previous work, we attempt to introduce reverse training to coreference resolution, as many coreferential relationships can be deemed as bidirectional, thus increasing the amount of useful information for model training.



\section{LLM-Based Coreference Resolution}
\label{sec:llm-cr}


\subsection{The Question-Answering (QA) Template}
A Question-Answering (QA) Prompt Template is a structured format that guides LLMs in generating accurate responses to questions. 
It follows the structure $Instruction+Sentences+Question$  for coreference resolution. The first part $Instruction$ guides the LLM in resolving candidate mentions. The second part $Sentences$ provides the necessary context in a set of sentences, including the one containing the target mention. The last part $Question$ specifies the target mention to analyze to find its coreferential mention. An example of the QA Template can be seen in Figure~\ref{fig:QA-document-template-example}.


\subsection{The Document Template}
The Document Template combines an instruction with a document for processing. The document marks candidate mentions and designates character positions where the LLM must output cluster IDs for these mentions.
These cluster IDs group coreferential mentions together, with all mentions in a cluster referring to the same entity. This structure enables the LLMs to perform coreference resolution. An example of the Document Template can be seen in Figure~\ref{fig:QA-document-template-example}.


\subsection{Supervised Fine-Tuning for Coreference Resolution}
To improve LLMs' performance in coreference resolution, we converted existing datasets into the two template formats described above for supervised fine-tuning (SFT). This process adapted open-source LLMs to the coreference resolution task.

Using the QA Template as an example, given a coreference dataset $D$, each document $d$ contains multiple mentions $m_k, k=1,2, \ldots, n$. Each queried mention $m_k$ has its corresponding resolved mention $r_k$ as the answer. By inserting $d$ and $m_k$ into the QA Template, we create a prompt $p_i$ and form a training data pair $(p_i, rp_i)$, where $rp_i$ equals $r_k$. This process transforms the dataset $D$ into a training dataset $T={(p_i, rp_i)}$. Using this training dataset $T$ and a large language model $M$, SFT aims to minimize the loss to get the optimal $M^*$:

\begin{equation}
\min_{M} \Sigma^{T}_{i=1} L(M(pi), rp_i)
\end{equation}

After fine-tuning, we use the optimal model $M^*$ for inference. The inference data will also be converted into the corresponding template format.

\begin{figure*}[h]
  \includegraphics[width=0.9\linewidth]{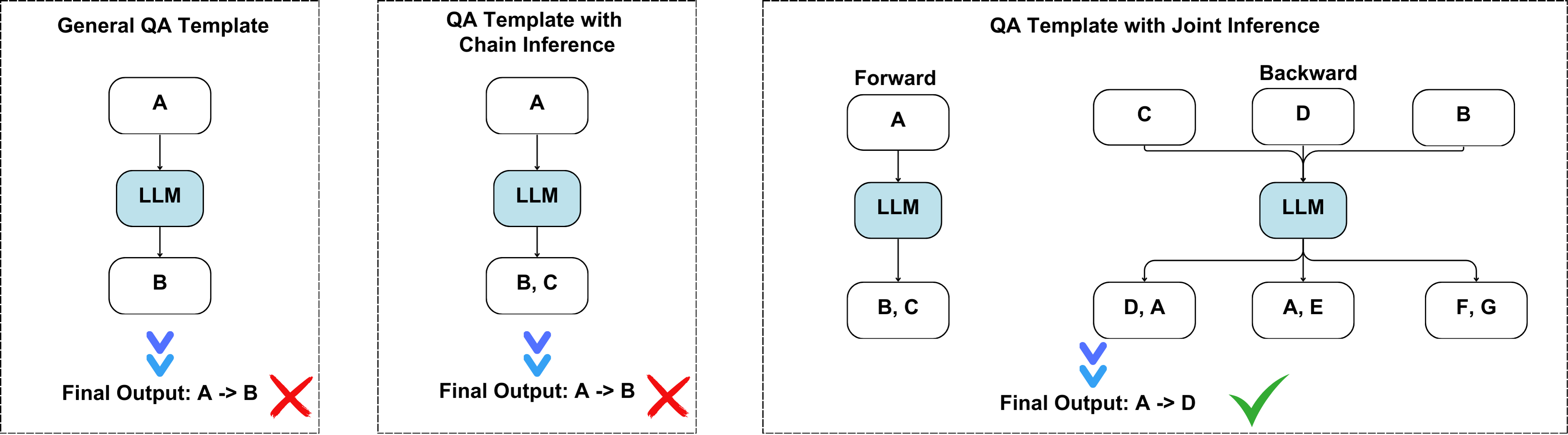}
  \caption{A comparison between different inference approaches.}
  \label{fig:joint-inference}
\end{figure*}

\section{The Refined QA Template Approach}
\label{sec:qa-template}

\subsection{Reversed Training for Conference Resolution}
\label{subsec:reversed-training}
Based on the previous work, we reverse the pattern ``A refers to B'' in the QA Template (input A, output B) to ``B refers to A'' (input B, output A). We propose two training modes: forward training and backward training. In this context, forward training is exemplified in Figure~\ref{fig:QA-document-template-example}. In the backward mode, however, as the position of the coreferential mention is uncertain, the backward prompt requires the entire document to be provided.

However, unlike other tasks, forward and backward training alone are not sufficient for the CR task, because it also involves singletons that do not refer to any other mentions. To address this issue, we designed a new question task specifically to identify mentions without antecedents: 

\begin{quote}
``List all phrases in the last sentence that do not refer to any previous phrases preceding them. Your response should be formatted as a Python list containing all phrases from the candidate list that do not refer to any preceding phrases. If no such phrases can be found, simply return `None.'''
\end{quote}

In summary, we designed three different QA tasks for reversed training: forward training, backward training, and finding all singletons.






\subsection{Learning to Generate a Chain}
As discussed in Section~\ref{subsec:reversed-training}, one may assume a canonical case that the target mention refers to only one coreferential mention, be it the antecedent in the forward variation, or the anaphor in the backward variation.
Actually, for a target mention, the number of coreferential mentions generated by LLM can be more than one ($n, n=0,1,2,...,i$). 
Specially, when $n=0$, the target mention is a singleton that fails to form a chain; when $n=1$, the target mention and its coreferential mention form a coreference chain with the minimal length. 
Prompting LLMs to generate multiple coreferential mentions, if any, may provide more information that model training may benefit from.
Thus, we advocate the method of learning to generate a coreference chain, rather than a single mention.
In practice, we train the LLM to identify the two most recent mentions, as shown in the second and third panels of Figure~\ref{fig:joint-inference}.

\subsection{Joint Inference}

Joint Inference is based on reversed training and learning to generate a chain. 
After the model generates results for forward, backward, and singleton tasks, we compare the multiple outputs to obtain the final correct answer.
Figure~\ref{fig:joint-inference} illustrates an example of Joint Inference. In this case, assuming the LLM incorrectly infers that A refers to B, we can correct this mistake by comparing the outputs of backward tasks C, D, and B, ultimately correcting that A refers to D.


The joint inference algorithm takes two inputs: a set of forward pairs $FP$ and a set of backward pairs $BP$. The algorithm first assigns initial weights to pairs in both groups, then updates these weights and stores the sum of weights for pairs appearing in both $BP$ and $FP$ in $W$. Pairs with weights exceeding 2 are used to create longer chains $C$, which are considered correct. Finally, the algorithm combines the pair weights $W$ and chains to generate the final output $(anaphor, antecedent)$. 

How to combine the pair weights $W$ and chains $C$ to generate the final output? The process iterates through all mentions in the weight list $W$. For each mention $a$, if any candidate referent $b, b \in W[a]$ has a weight $W[a][b] \geq 2$, the referent of $a$ is considered found. Otherwise, the algorithm examines pairs of candidate referents $b, b \in W[a]$ and $d, d \in W[a]$. If both $b$ and $d$ co-occur in any chain $C$, their respective weights in $W$ are incremented by 1. This process updates all pair weights using the chain information. After updating is complete, the algorithm selects the referent phrase with the highest weight for each mention to create the final pair $(anaphor, antecedent)$.







\begin{figure}[h]
  \includegraphics[width=\columnwidth]{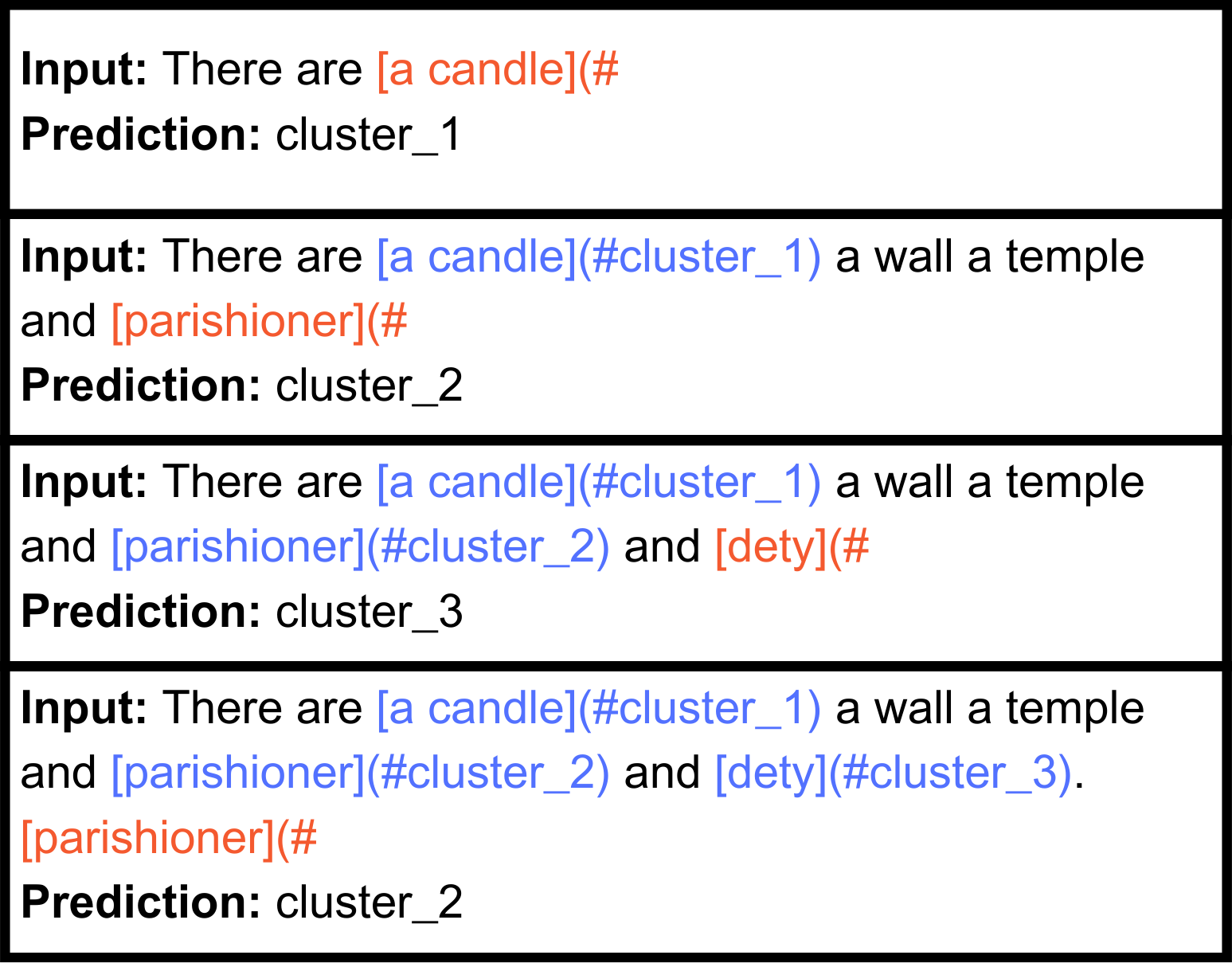}
  \caption{An example of the iterative document generation process of the refined Document Template. For each iteration, the input consists of a text segment containing a marked mention and its designated position at the end. The LLMs then predict the cluster ID for the marked mention. This predicted ID and its following content are appended to the end, creating a new prompt for analyzing the next mention. This sequential process continues till all mentions have been analyzed.}
  \label{fig:iterative-generation}
  \vspace{-1em}
\end{figure}

\section{Iterative Document Generation}
\label{sec:Document-Template}

\citet{le2023largelanguagemodelsrobust}'s Document Template provides the complete document with marked candidate mentions and designated character positions. The LLMs must input cluster IDs for these candidate mentions simultaneously at the designated positions. However, we discovered that processing the entire document and analyzing all mentions at once led to hallucination—where the LLM would generate irrelevant words not present in the document instead of completing the intended task. This issue significantly reduced the model's UA-SA score from approximately 89 (when excluding hallucination results from evaluation) to 59 (when including hallucination results). 

To address this issue, we modified the full document generation process to generate only one mention's cluster ID at a time, as shown in Figure~\ref{fig:iterative-generation}. The advantage of generating just one cluster ID per step is that it significantly reduces the difficulty of content generation for the LLM, helping to avoid hallucinations.
The process begins with a text segment containing the document's first marked mention and its designated position. The LLM assigns a cluster ID to this mention. After filling this predicted cluster ID into the designated position, a new text segment is created for analyzing the next mention. This sequential analysis continues until all mentions in the document are processed. During training, the cluster IDs are filled with correct values. During inference, they are filled with IDs generated by the LLMs. 

In prompt design, since only a cluster ID is predicted, using just the partial sentence from Figure~\ref{fig:iterative-generation} may not be enough for accurate prediction. 
Considering the context is crucial for understanding the coreference relationship of the mention. 
Therefore, based on the Document Template in Figure~\ref{fig:QA-document-template-example}, we removed the full text from the input and instead added the relevant sentence containing the last mention. 
Unlike in Figure~\ref{fig:QA-document-template-example}, this relevant sentence does not include the mention's position annotation.


\section{Experiment}
\label{sec:experiment}

\begin{table*}[t]
    \centering
    \resizebox{1.9\columnwidth}{!}{
    \smallskip\begin{tabular}{l|cc|cc|cc|cc}
        \hline
         & \multicolumn{2}{c|}{LIGHT} &  \multicolumn{2}{c|}{Penn Treebank}
         &  \multicolumn{2}{c|}{TRAINS} & \multicolumn{2}{c}{OntoNotes} \\
        \hline
       \bf Approach & \bf UA & \bf -S -SA & \bf UA & \bf -S -SA & \bf UA & \bf -S -SA & \bf UA & \bf -S -SA \\
       \hline \hline

LLAMA $\mathbb{S}^{\dagger}$ & 84.02  &  70.39 & 83.83 & 70.60 & 78.63  & 75.03 & 82.86 & 84.05 \\

LLAMA $\mathbb{R}$ &  
85.77  & 73.20 & 86.86   &  77.12 & 79.20 & 74.68  & 81.23 & 82.58 \\
\hline

LLAMA $\mathbb{SL}$ & 86.64  &  75.69 & 86.09 & 74.47 & 79.58 & 75.07 & 81.80  & 82.83 \\
LLAMA $\mathbb{RL}$&  88.20 & 77.81
& $\textbf{87.99}$ & $\textbf{78.56}$ & 81.76 & 78.36  & $\textbf{83.04}$ & $\textbf{84.60}$  \\
\hline

LLAMA $\mathbb{SC}$ & 87.25 & 75.54 & 85.98 & 74.90 & 78.18 & 73.38 &  &  \\

LLAMA $\mathbb{RC}$ &  87.30 &  76.18 &   85.78& 74.84 &  81.97 & 77.92 &  &  \\

\hline

LLAMA $\mathbb{SCL}$ & 87.20 & 77.09 & 85.46 & 74.25 & 82.69 & 78.37 &  &  \\

LLAMA $\mathbb{RCL}$&  $\textbf{89.29}$ &  $\textbf{80.40}$ & 87.46& 77.86 &  $\textbf{83.67}$ & $\textbf{81.02}$ &  &  \\

\hline
QWEN $\mathbb{SL}$ &  82.66 & 68.35
& 85.72 & 73.66 & 80.79 & 76.48  &   \\
QWEN $\mathbb{RL}$ & $\textbf{87.08}$  &  $\textbf{76.49}$ & $\textbf{86.19}$ & $\textbf{75.10}$ & $\textbf{82.62}$ & $\textbf{79.64}$ & 
\\
        \hline 

QWEN $\mathbb{S}^{\dagger}$ &  82.48 & 67.50
& 84.22 &  71.54 & 78.14 & 72.48  &   \\
QWEN $\mathbb{R}$ & 86.35  &   75.19 & 85.98 &  74.93 & 79.54 &  74.59 & 
\\
        \hline 
    \end{tabular}
    }
    \caption{Comparison of the training effects between Reversed Learning and fine-tuning based on original data. $\dagger$ denotes results reproduced from \citet{gan-etal-2024-assessing}. 
    Our reproduced results outperform those reported by \citet{gan-etal-2024-assessing} for two main reasons:
(1) we fixed an evaluation issue in their original setup, where non-referring expressions were not excluded during scoring; and
(2) our model architecture differs from theirs, which also contributes to the performance gains.
    To facilitate a lightweight comparison, we conducted testing on a randomly selected subset of the test dataset only, which is sufficient to verify the effectiveness of our method.}
    \label{table:reversed-total}
    \vspace{-0.5em}
\end{table*}

\subsection{Experimental Setup}
\paragraph{Dataset.}
We used the same datasets as those of ~\citet{gan-etal-2024-assessing} and \citet{le2023largelanguagemodelsrobust}: OntoNotes 5.0 \citep{pradhan-etal-2012-conll} and {\CRACS} \cite{poesio-et-al:CRAC18,yu-et-al:CODI-CRAC-22}. 
We utilize three subsets of {\CRACS}: LIGHT, Penn Treebank, and TRAINS.
The training set sizes vary significantly across datasets, with OntoNotes having the largest set, while LIGHT, Penn Treebank, and TRAINS comprise 2.73\%, 24.95\%, and 10.21\% of OntoNotes' training set size, respectively.
Consistent with the experimental setup of previous work, we used gold mentions. It should be noted that the QA Template's prompt does not specify the exact mention in the question, making it slightly more challenging compared to the Document Template.

To ensure consistency with prior work, we use a subset of the test set for comparison with~\citet{gan-etal-2024-assessing}, and the full test set for comparison with~\citet{le2023largelanguagemodelsrobust}.
The four datasets used in this study are shown as follows:

\begin{itemize}[leftmargin=*,noitemsep,topsep=0em]
    \item $\textbf{OntoNotes}$: We extracted 2,516 training documents from its training set, resulting in 128,319 training examples for the QA Template. We randomly selected 55 test documents from its test set, yielding 3,624 test examples for the QA Template.
    
  \item $\textbf{LIGHT}$: We used the full LIGHT training set, resulting in 3,499 training examples for the QA Template, and randomly selected 4 test documents, yielding 872 test examples.
    
      \item $\textbf{Penn Treebank}$: We used the full Penn Treebank training set, resulting in 31,979 training examples for the QA Template, and randomly selected 5 test documents, yielding 1067 test examples.
      
  \item $\textbf{TRAINS}$: We used the full Penn Treebank training set, resulting in 13,087 training examples for the QA Template, and randomly selected 4 test documents, yielding 757 test examples.

  \item $\textbf{Combination}$: We combined the training sets from LIGHT, Penn Treebank, and TRAINS. No test set was used.

\end{itemize}

\paragraph{Models.}
To validate the feasibility of our method, we fine-tuned two different large language models (LLMs) as base models: LLama 3.1~\citep{grattafiori2024llama3herdmodels} and Qwen 2.5~\cite{qwen2025qwen25technicalreport}. 
All experiments were conducted using LLAMA Factory \citep{zheng2024llamafactory}, with LoRA-based~\citep{hu2021loralowrankadaptationlarge}  fine-tuning utilizing the bf16 precision. The relevant hyperparameters were configured as follows:
\begin{itemize}[leftmargin=*,noitemsep,topsep=0em]
\item preprocessing\_num\_workers: 16
\item per\_device\_train\_batch\_size: 1
\item gradient\_accumulation\_steps: 4
\item learning\_rate: 1.0e-4
\item num\_train\_epochs: 3.0
\item lr\_scheduler\_type: cosine
\item warmup\_ratio: 0.1
\end{itemize}

Other parameters were set to the default values provided by LLAMA Factory \citep{zheng2024llamafactory}. The experiments were conducted on single or multiple A100 GPUs (40GB/80GB). For multi-GPU setups, DeepSpeed was used to optimize parallel training.

The symbols associated with the models in this section are defined as follows:
\begin{itemize}[leftmargin=*,noitemsep,topsep=0em]
\item $\textbf{LLAMA}$ The LLama 3.1 8B Instruct model.
\item $\textbf{QWEN}$ The Qwen 2.5 7B Instruct model.
\item $\mathbb{R}$ indicates that the model is trained using reversed learning based on the QA Template.
\item $\mathbb{S}$ indicates that the model is trained using standard fine-tuning based on the QA Template.
\item $\mathbb{C}$ represents that the model is trained on the Combination set.
\item $\mathbb{OC}$ indicates that the model is trained on the OntoNotes, and then evaluated separately on {\CRACS} and OntoNotes.

\item $\mathbb{L}$ indicates that the model is trained to predict a coreference chain (i.e., two mentions instead of just one) based on the QA Template. If the model also uses reversed learning (denoted by $\mathbb{RL}$), joint inference will automatically be applied in this case.

\item $\mathbb{F}$ indicates that the model is trained to predict a full document based on the Document Template.

\item $\mathbb{I}$ indicates that the model is trained using the iterative document generation method based on the Document Template.

\end{itemize}

\paragraph{Metrics.}
For the QA Template, we use the Universal Anaphora (UA) scorer \cite{poesio-etal-2024-universal} to evaluate different approaches, where UA represents the
full score, and `-S -SA' indicate the CoNLL score (i.e., without singletons and split antecedents). 
For the Document Template, consistent with the experiment setup of ~\citet{le2023largelanguagemodelsrobust}\footnote{The Reference Coreference Scorer \cite{pradhan-EtAl:2014:P14-2} does not score singletons (-S -SA). However, \citet{le2023largelanguagemodelsrobust} retrained singletons in their evaluation. This may not be a big problem when evaluating on corpora without singleton mentions (e.g., OntoNotes), as singletons will only appear in system clusters. However, in corpora with singleton mentions (e.g., {\CRACS}), this difference in setting can result in a change of more than 10\% in CoNLL $F_1$. To make a fair comparison, we followed the same settings when compared to their system. }, we use `-SA' to indicate UA without split antecedents for evaluation and comparison across different methods. We also calculate the Pass score and Exact Match (EM) score to quantify the extent of hallucinations.
“Pass” denotes the percentage of documents that passed the alignment check as defined in \citet{le2023largelanguagemodelsrobust}'s work. “EM” refers to the exact match rate between the generated text and the original reference text.

\paragraph{Baselines.} 
For both the QA and the Document Templates, we adopted the approaches proposed in \citet{le2023largelanguagemodelsrobust}'s work as our baseline methods. However, their experiments were conducted using LLama 2, which is now somehow outdated. Recently, the LLama 3.1 8B and Qwen 2.5 have been released, featuring architectural improvements and training on a significantly larger dataset, resulting in noticeably enhanced performance. 

To establish updated baseline results for fair comparison, we fine-tuned LLama 3.1 8B and Qwen 2.5 using the same methods described by ~\citet{le2023largelanguagemodelsrobust}. In our experiments, the fine-tuned models are denoted with $\mathbb{S}$ for the QA Template and $\mathbb{F}$ for the Document Template.

\begin{table}[t]
    \centering
    \resizebox{0.99\columnwidth}{!}{
    \smallskip\begin{tabular}{l|cc|cc|cc}
        \hline
         & \multicolumn{2}{c|}{LIGHT} &  \multicolumn{2}{c|}{Penn Treebank}
         &  \multicolumn{2}{c}{TRAINS} \\
        \hline
       \bf Approach & \bf UA & \bf -S -SA & \bf UA & \bf -S -SA & \bf UA & \bf -S -SA  \\
       \hline \hline

LLAMA $\mathbb{RL}$ &  88.20 & 77.81
& 87.99 & 78.56 & 81.76 & 78.36    \\

w/o Join Inference &  87.95  & 77.74
& 87.56 & 77.83 & 81.45 & 77.46    \\

w/o $\mathbb{R}$ & 86.64  &  75.69 & 86.09 & 74.47 & 79.58 & 75.07  \\

w/o $\mathbb{L}$  & 84.02  &  70.39 & 83.83 & 70.60 & 78.63  & 75.03   \\

\hline

LLAMA $\mathbb{RCL}$&  89.29 &  80.40 &   87.46& 77.86 &  83.67 & 81.02  \\
w/o Join Inference &  88.46 &  78.87 &   86.79& 76.55 &  83.08 & 79.88  \\

w/o $\mathbb{R}$ & 87.20 & 77.09 & 85.46 & 74.25 & 82.69 & 78.37    \\

w/o $\mathbb{L}$ & 87.25 & 75.54 & 85.98 & 74.90 & 78.18 & 73.38  \\

        \hline 

QWEN $\mathbb{RL}$ & 87.08  &   76.49 & 86.19 & 75.10 & 82.62 & 79.64  \\
w/o Join Inference &  86.59 & 76.66 & 85.99 & 74.46 & 81.13 & 77.52
\\
w/o $\mathbb{R}$ & 82.66 & 68.35
& 85.72 & 73.66 & 80.79 & 76.48    \\

w/o $\mathbb{L}$ &  82.48 & 67.50
& 84.22 &  71.54 & 78.14 & 72.48  
   \\
    \end{tabular}
    }
    \caption{Ablation Analysis of Different QA Template Approaches.}
    \label{table:Ablation study}
    \vspace{-1em}
\end{table}



\begin{table*}[t]
    \centering
    \resizebox{1.9\columnwidth}{!}{
    \smallskip\begin{tabular}{l|ccc|ccc|ccc|ccc}
        \hline
         & \multicolumn{3}{c|}{LIGHT} &  \multicolumn{3}{c|}{Penn Treebank}
         &  \multicolumn{3}{c|}{TRAINS} & \multicolumn{3}{c}{OntoNotes} \\
        \hline
       \bf Approach & \bf   -SA & \bf Pass &  \bf EM & \bf \bf  -SA & \bf Pass & \bf  EM &\bf  \bf  -SA & \bf Pass & \bf  EM & \bf  \bf  -SA & \bf Pass& \bf  EM  \\
       \hline \hline

\hline 
LLAMA $\mathbb{OCF}^{\dagger}$ & 86.6 & 92.11\% &81.58\%& 82.5 & 100\% & 70\% & 73.4 & 100\%  &62.50\%& $\textbf{95.3}$ & 99.71\%& 97.41\% \\
LLAMA $\mathbb{OCI}$ & $\textbf{87.8}$ & 100\%  &100\% & $\textbf{85.2}$ & 100\%  &100\% & $\textbf{76.4}$ & 100\%  &100\% & 94.9 & 100\%  &100\%\\

\hline 
QWEN $\mathbb{OCF}^{\dagger}$ & $\textbf{87.0}$ & 97.37\% & 92.11\% & $\textbf{83.2}$ & 98.33\% &71.67\% & 77.2 & 100\% & 81.25\%& $\textbf{95.0}$ & 99.14\% & 97.13\%\\
QWEN $\mathbb{OCI}$ & 85.7 & 100\%  &100\% & 80.6 &100\%  & 100\% & $\textbf{79.6}$ &100\%  &100\% &94.8 &100\%  & 100\%\\
        \hline
    \end{tabular}
    }
    \caption{Comparison of Training Outcomes Between Iterative and Full Document Generation on the full test dataset.
 $\dagger$ denotes results reproduced from \citet{le2023largelanguagemodelsrobust}.
 Tables~\ref{table:reversed-total} and ~\ref{table:Ablation study} are based on a subset of the test set, whereas this table uses the full test set to ensure a fair comparison with \citet{le2023largelanguagemodelsrobust}, which uses the complete test set.
}
    \label{table:doc-total}
    \vspace{-1em}
\end{table*}


\subsection{QA Template Results}
From the results shown in Table \ref{table:reversed-total}, we observe that models trained with reversed learning generally outperformed those trained using standard fine-tuning across most datasets, in terms of both the full score (UA) and the CoNLL score excluding singletons and split antecedents (denoted as -S -SA). 
This held true for both the LLama 3.1 and Qwen 2.5 models.

For both LLMs, reversed learning consistently yielded higher performance across all datasets, except for the comparison between LLAMA $\mathbb{S}$ and LLAMA $\mathbb{R}$ on the OntoNotes dataset.
This suggests that when the training dataset is sufficiently large (128,319 training examples in OntoNotes) and diverse, it is enough for the models to learn the task.
After adding Joint inference, the performance of LLAMA $\mathbb{RL}$ successfully surpassed that of LLAMA $\mathbb{SL}$ and LLAMA $\mathbb{S}$. This indicates that the method proposed in this paper indeed enhances the model's ability to solve CR under the QA Template.

On the other hand, for smaller datasets like LIGHT, Penn Treebank, TRAINS, and their combination (which vary in size from 3,499 to 48,565 examples), reversed learning consistently outperformed standard fine-tuning, highlighting its advantage in situations where training data is limited.
For Qwen 2.5, similar trends can be observed. Among all the comparisons, the largest improvement was seen when transitioning from QWEN $\mathbb{SL}$ to QWEN $\mathbb{RL}$. Specifically, the `UA' score increased by 4.56 (from 82.66 to 87.08) and the `-S -SA' score improved by 8.14 (from 68.35 to 76.49).

\paragraph{Ablation Study.}

To understand the contributions of each component, we conducted an ablation study by removing key elements from the model. We examined three components: Joint Inference, Reversed Learning ($\mathbb{R}$), and coreference chain prediction ($\mathbb{L}$). Results in Table \ref{table:Ablation study} show their impact on performance across the LIGHT, Penn Treebank, and TRAINS datasets.

It is clear that the best performance is achieved when all components are present. 
Table \ref{table:Ablation study} confirms that each component clearly contributes to the model's overall performance.
Removing any of these components leads to a reduction in performance, highlighting the effectiveness of the proposed approach in solving the coreference resolution task using the QA Template.













\subsection{Document Template Results}



From Table~\ref{table:doc-total}, we observe that hallucinations were severe in full document generation, as indicated by significantly lower EM scores and alignment check pass rates in the $\mathbb{OCF}$ setting. In contrast, our proposed iterative generation method ($\mathbb{OCI}$) not only eliminated hallucinations—achieving a 100\% pass rate on the alignment check across all datasets—but also significantly improved the Exact Match (EM) scores.

The EM metric provides a stricter evaluation of output fidelity, measuring whether the generated text exactly matches the gold reference. Our method achieved 100\% EM across all datasets, underscoring its ability to produce highly faithful outputs. In comparison, full document generation often failed to match reference texts exactly, with EM scores ranging from 62.50\% to 92.11\% depending on the dataset and model. This further supports our claim that iterative generation produces more precise and consistent outputs.

Moreover, even when focusing solely on documents that passed the alignment check, our method still consistently outperformed full generation in EM, revealing that alignment check alone may not capture finer-grained errors that EM can detect. This discrepancy is especially apparent in the TRAINS dataset. Although the alignment pass rate was 100\% for both $\mathbb{OCF}$ and $\mathbb{OCI}$, EM for $\mathbb{OCF}$ remained low (62.50\%–81.25\%), suggesting that its outputs, while structurally aligned, still contained subtle inconsistencies or content deviations.

These results collectively demonstrate that full document generation lacks robustness and struggles to achieve both structural alignment and content fidelity. Our iterative approach not only ensures alignment but also guarantees high-fidelity generation, as reflected in both pass rates and exact match accuracy.

\section{Discussion}
\subsection{Why does the Document Template Work Better?}
The Document Template excels because it explicitly marks all mentions, guiding the LLM to focus only on relevant parts of the text. This structured approach reduces the complexity, as the model does not need to identify appropriate mentions within the entire document, as it does with the QA Template. Essentially, it simplifies the task to something akin to a classification problem, where the LLM identifies coreference relations within predefined segments, making it easier and more accurate.

\subsection{Do We Still Need the QA Template?}
From a purely CR perspective, the Document Template is indeed better because the task is simpler and easier to train. However, the goal of CR research is not just to resolve coreferences but also to enhance the model's understanding of logical relationships in a language. While the Document Template performs better, it may only train a classification model that relies on mention annotations or detection, which limits its integration with other tasks. 
We consider it essentially a classification model because, in the Document Template task, the large language model only needs to generate the cluster ID to which a given mention belongs. In other words, it simply has to decide whether the mention belongs to a new cluster (a new cluster ID) or an existing one (an existing cluster ID).
In contrast, the QA Template requires no mention annotations in the prompt, and the trained model is likely to have a better understanding of language, making it easier to integrate with other tasks. Therefore, in the long run, learning from the QA Template remains irreplaceable, particularly for cross-task applications and improving language comprehension.

\section{Conclusion}
 In this paper, we have analyzed the limitations of current approaches to using LLMs for coreference resolution, particularly the QA and Document Templates, and then proposed two novel methods: Reversed Training with Joint Inference, and Iterative Document Generation. Our findings can be summarized as follows. First, Reversed Training enhances the QA Template method. Second, Iterative Document Generation eliminates hallucinations with an improvement in task performance.  These methods and techniques proposed in our study form an integrated solution to enhance model learning in LLM-based coreference resolution.


\section*{Limitations}
Although the methods proposed in this paper have been experimentally validated on English coreference resolution tasks, they have not been extensively tested on coreference tasks in other languages. Therefore, we are currently unable to assess the applicability of these methods to other languages. Additionally, due to limited computational resources, the methods have not been tested on larger-scale LLMs (e.g., models with 70B parameters or more), so we are uncertain about their performance of such models.

\section*{Acknowledgments}
We thank the anonymous reviewers for their helpful comments.
Yujian Gan, Juntao Yu and Massimo Poesio acknowledge financial support from the UK EPSRC under grant EP/W001632/1.

\bibliography{custom,  yg, lyn}

\appendix



\end{document}